\documentclass[10pt,twocolumn,letterpaper]{article}

\usepackage[pagenumbers]{iccv} 

\definecolor{iccvblue}{rgb}{0.21,0.49,0.74}
\usepackage[pagebackref,breaklinks,colorlinks,allcolors=iccvblue]{hyperref}

\def\paperID{0} 
\def\confName{ICCV}
\def\confYear{2025}

\title{The Escalator Problem: Identifying Implicit Motion Blindness in AI for Accessibility}

\author{Xiantao Zhang\\
    Beihang University\\
    {\tt\small zhangxiantao@buaa.edu.cn}
}

\usepackage{CJK}
\usepackage{amsfonts}
\usepackage{multirow}
\usepackage{enumitem}
\usepackage{layout}
\usepackage{wrapfig}
\usepackage{caption}
\definecolor{LightCyan}{RGB}{224,255,255}
\usepackage{listings}
\usepackage{tabularx}
\usepackage{array}

\usepackage{lipsum} 

\begin{document}
\maketitle
\begin{abstract}

Multimodal Large Language Models (MLLMs) hold immense promise as assistive technologies for the blind and visually impaired (BVI) community.
However, we identify a critical failure mode that undermines their trustworthiness in real-world applications.
We introduce the \textbf{\textit{Escalator Problem}}---the inability of state-of-the-art models to perceive an escalator's direction of travel---as a canonical example of a deeper limitation we term \textbf{\textit{Implicit Motion Blindness}}.
This blindness stems from the dominant frame-sampling paradigm in video understanding, which, by treating videos as discrete sequences of static images, fundamentally struggles to perceive continuous, low-signal motion.
As a position paper, our contribution is not a new model but rather to: (I) formally articulate this blind spot, (II) analyze its implications for user trust, and (III) issue a call to action.
We advocate for a paradigm shift from purely semantic recognition towards robust physical perception and urge the development of new, human-centered benchmarks that prioritize safety, reliability, and the genuine needs of users in dynamic environments.

\end{abstract}    
\section{Introduction}
\label{sec:intro}

\begin{figure}[t]
    \centering
    \includegraphics[width=\linewidth]{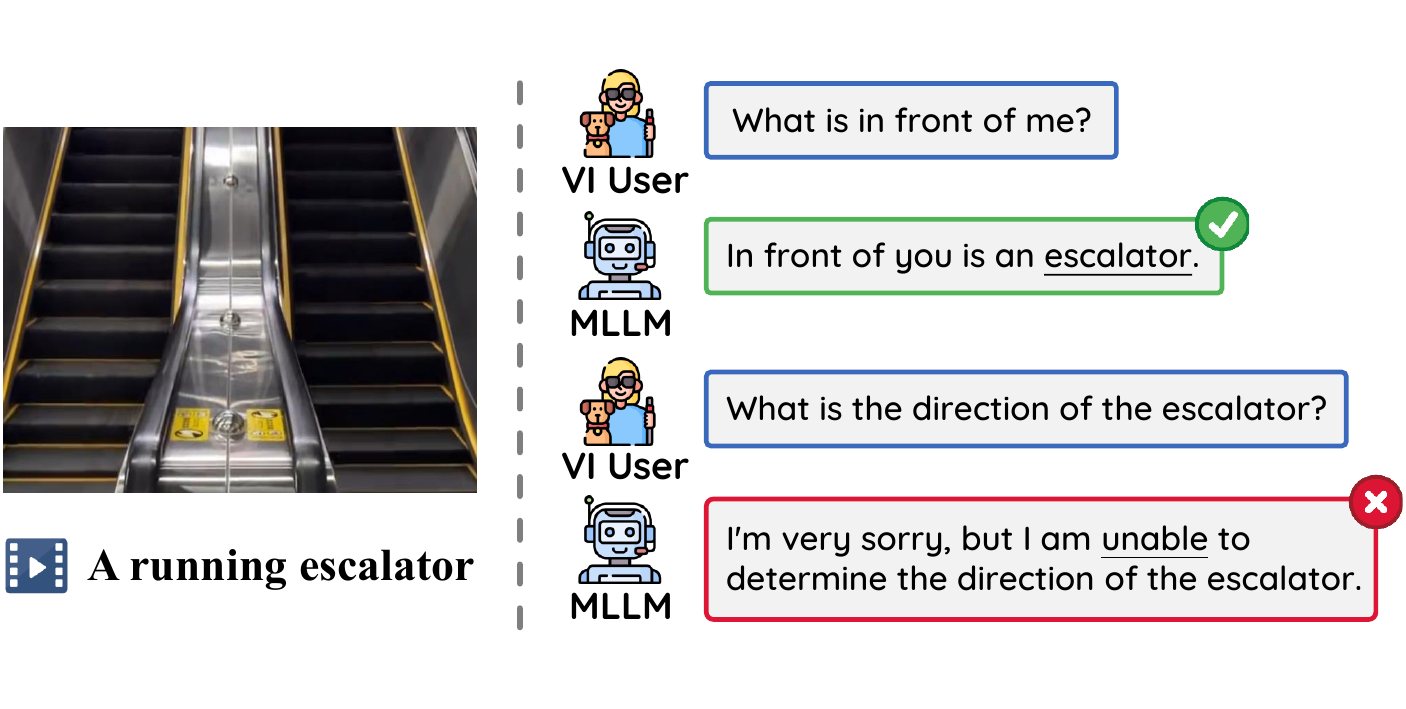}
    
    \caption{The \textit{Escalator Problem}: A Real-World Failure of Motion Perception. An interaction with the MLLM demonstrates a critical failure. While the model correctly identifies the object as an escalator, it is unable to determine its direction of motion, a vital piece of information for a BVI user's safe navigation.
    This highlights the concept of \textit{Implicit Motion Blindness}.
    }
    \label{fig:motivation}
    \vskip -0.2in
\end{figure}

Vision foundation models (VFMs) and Multimodal Large Language Models (MLLMs) have ushered in an era of unprecedented progress in artificial intelligence, with the potential to revolutionize assistive technologies.
For the blind and visually impaired (BVI) community, these advancements promise a future where AI can act as a ``visual interpreter,'' describing complex scenes, identifying objects, and empowering greater independence in daily life. This vision aligns perfectly with the core mission of the accessibility research community: to harness cutting-edge technology to dismantle barriers and foster a more inclusive world.
The prospect of MLLMs acting as real-time, conversational ``visual interpreters'' via live video is no longer science fiction but a rapidly commercializing reality, as exemplified by models like GPT-4o \cite{openai2024gpt4ocard} and Qwen2.5-Omni \cite{xu2025qwen25omni}.
These systems, capable of fluid dialogue about a user's live surroundings, represent an exciting frontier.

However, despite this encouraging progress, a critical gap persists between current model capabilities and the demands of real-world navigation.
In this paper, we identify a fundamental failure mode using the \textit{Escalator Problem} as a canonical, illustrative case: determining the direction of a moving escalator from a first-person video. To demonstrate this blind spot, we presented video footage of a standard escalator to leading MLLMs. Their responses were consistently alarming; as illustrated in Figure~\ref{fig:motivation}, even top-tier models often fail this seemingly trivial task. They tend to default to describing a static scene or explicitly state their inability to perceive motion. This simple observation forms the crux of our investigation.

This paper posits that this failure is not an isolated glitch but rather a symptom of a deeper, systemic issue we term \textit{Implicit Motion Blindness}.
We argue that the dominant frame-sampling paradigm in video understanding—which processes videos as a sequence of discrete, static images—is ill-equipped to perceive low-signal, continuous motion.
The crucial directional information in the escalator video does not reside within any single frame, but exists implicitly between them, in the subtle, continuous flow that is effortlessly perceived by the human visual system. The models' inability to capture this flow reveals a critical blind spot for tasks vital for safe and independent navigation.

As a position paper, our contribution is not a new model or dataset but rather the identification and articulation of this critical blind spot to steer the community's focus. We make the following three contributions:

\begin{enumerate}[label=\Roman*]
    \item We identify and formally articulate the \textit{Escalator Problem} as a canonical, real-world failure case that highlights the limitations of current MLLMs in perceiving continuous, low-signal motion.
    \item We analyze the broader implications of this \textit{Implicit Motion Blindness} for the BVI community, arguing that it represents a significant barrier to building trustworthy assistive technologies and can erode user trust in real-world deployments.
    \item We issue a call to action for the research community to shift its focus from purely semantic recognition towards robust physical perception, and call for the co-development of new, human-centric evaluation paradigms that prioritize safety, reliability, and trust in assistive AI.
\end{enumerate}

\begin{figure*}[ht]
    \centering
    \includegraphics[width=\linewidth]{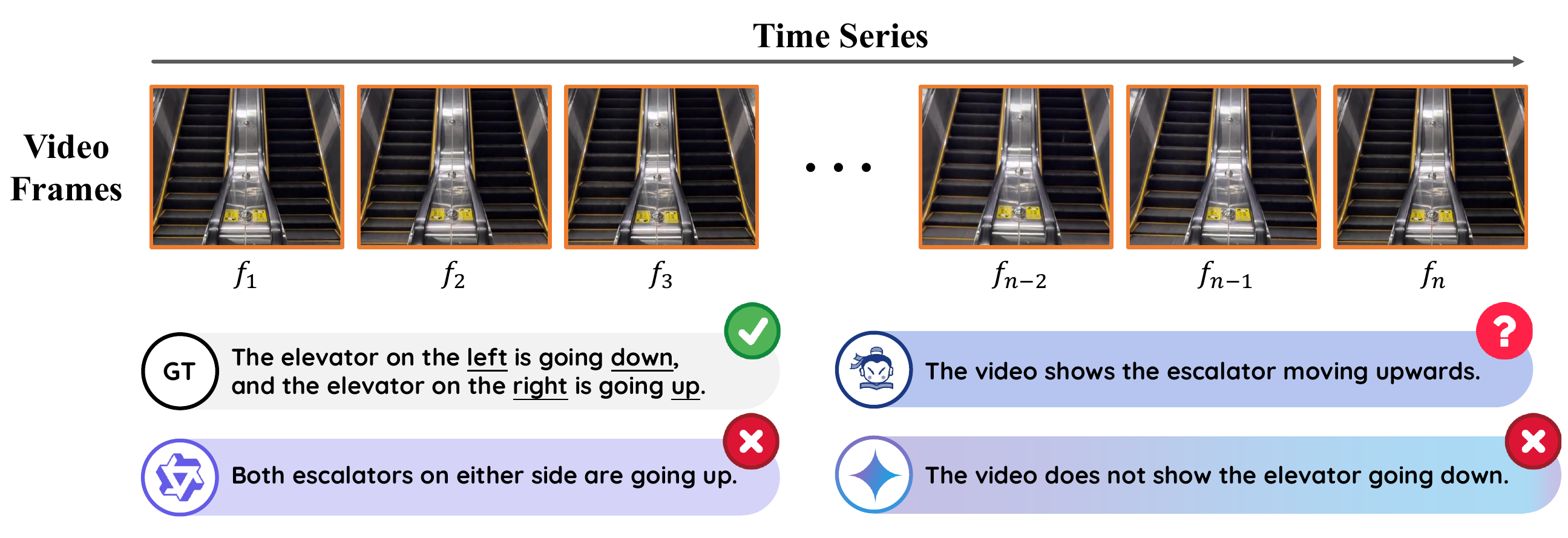}
    
    \caption{An Illustrative Example of \textit{Implicit Motion Blindness}. When presented with a video of two adjacent escalators, current MLLMs fail to correctly describe the motion. The Ground Truth (GT) indicates opposite directions, but model outputs are incorrect, either guessing the wrong direction or denying the presence of motion altogether. This empirically validates the limitations of the current frame-sampling paradigm for tasks requiring fine-grained temporal understanding.}
    \label{fig:evidence_of_failure}
    \vskip -0.1in
\end{figure*}

\section{Background and Related Work}
\label{sec:related_work}

\subsection{Video Perception in MLLMs}
\label{subsec:video_perception_in_mllms}

The primary method for MLLMs to process video is through \textbf{sparse frame sampling}. This technique involves selecting a small, discrete subset of frames to represent the entire video sequence, drastically reducing the number of visual tokens for the model to process. Common methods include uniform sampling and frames-per-second (FPS) sampling \cite{lin2024videollava, wang2024qwen2vl, bai2025qwen25vl, wang2024internvideo2, wang2025internvideo25}.

While computationally efficient, this approach is fundamentally lossy. By discarding most frames, sparse sampling severs the continuous dynamic information that constitutes motion and temporal progression. Consequently, models can miss crucial events and struggle with tasks requiring fine-grained spatiotemporal understanding \cite{shi2025mavors}. For a typical 5-minute video, a model limited to 64 frames would sample at a rate of one frame every five seconds, making detailed motion analysis nearly impossible \cite{hong2025motionbench}. This method treats a video as an unordered ``bag of frames,'' sacrificing temporal integrity for manageable data size.

\subsection{Video Understanding Benchmarks}
\label{subsec:video_bench}

Leading video understanding benchmarks, such as Kinetics \cite{kay2017kinetics} and ActivityNet \cite{ghanem2018activitynet}, have been pivotal in advancing research.
These datasets primarily focus on \textbf{explicit action recognition}, where models are tasked with classifying activities like ``running'' or ``playing guitar.''

However, these benchmarks often have a ``\textit{static appearance bias}'' \cite{salehi2024actionatlas}. Many actions can be identified from a single frame, such as classifying diving by seeing a pool, often making temporal information redundant \cite{zhou2018temporalrelationalreasoningvideos}. This is confirmed by experiments where models perform well even with shuffled frames, showing that true temporal understanding is not always being evaluated \cite{feng2025breakingvideollmbenchmarks}. Consequently, these benchmarks test for high-level action classification where visual clues are often in a few keyframes, rather than assessing a model's ability to reason about events over time.

\subsection{AI for Accessibility}
\label{subsec:ai_for_accessibility}

AI has significantly advanced assistive technology for the visually impaired, with a global population of over 2.2 billion \cite{naayini2025aipoweredassistivetechnologiesvisual}. Applications like Microsoft's Seeing AI \cite{microsoft2017seeingai} and Google Lookout \cite{google2019lookout} excel at static perception tasks, including optical character recognition (OCR), product identification, and scene description. Wearable devices such as Envision Glasses \cite{envision2020envisionglasses} and the NOA vest \cite{biped2024noa} offer more continuous navigation support by providing obstacle detection and GPS navigation.

Despite these advances, a critical gap remains in handling dynamic, unpredictable environments \cite{silva2025navigationframeworkblindvisually}. Most assistive technologies focus on obstacle detection and static scene description, falling short of the nuanced understanding required for safe mobility \cite{naayini2025aipoweredassistivetechnologiesvisual}. They can identify an ``escalator'' but struggle to interpret its dynamic context, such as whether it is broken and must be walked on. This disconnect between low-level perception and high-level situational awareness is a major barrier to true navigational independence and safety.

\section{Why Models Can't See What's Moving}
\label{sec:failure}

The failure of sophisticated MLLMs on the Escalator Test is not a matter of insufficient parameters or inadequate training data in the conventional sense. Rather, it exposes a fundamental misalignment between the strategy models use to ``see'' and the nature of visual information in the physical world. To understand the failure, we must first appreciate the profound difference between how humans and current AI models perceive continuous motion.

\subsection{The Human Advantage: Perceiving the Flow}

For a human observer, determining the escalator's direction is a pre-attentive, almost instantaneous process. We do not analyze the scene as a series of static photographs. Instead, our visual system is exquisitely tuned to perceive \textit{optical flow}—the pattern of apparent motion of objects, surfaces, and edges in a visual scene caused by the relative motion between an observer and the scene. We perceive a continuous stream of information. Our brain effortlessly detects the subtle relative movements: the handrail gliding smoothly past a stationary pillar, the seamless cascade of the steps, which creates a waterfall effect, and the gentle parallax shift of background elements relative to the foreground.

This perception is holistic and relational.
The directional signal is not contained in the metallic texture of a single step but in the collective, coherent movement of all steps relative to the static surroundings. It is this perception of flow, not the recognition of objects, that allows for immediate and confident judgment.
This holistic, physics-aware process stands in stark contrast to the frame-by-frame object identification paradigm of MLLMs.
The visual evidence is clear and unambiguous precisely because our brains are built to process temporal dynamics as a primary source of information about the world.

\subsection{The Model's Blind Spot: Frame-Level Myopia}

\begin{figure*}[ht]
    \centering
    \includegraphics[width=\linewidth]{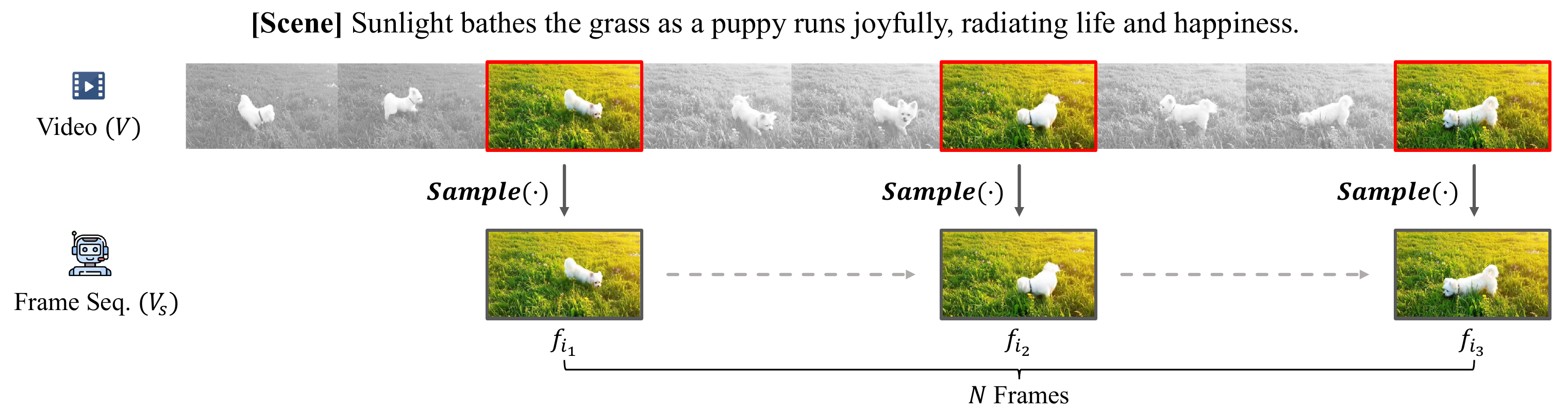}
    
    \caption{The Frame-Sampling Paradigm as the Source of \textit{Motion Blindness}. This diagram illustrates how MLLMs process video by sampling a sparse sequence of frames ($V_s$) from the original video ($V$). For high-level semantic descriptions, this is often sufficient. However, for continuous, low-signal motion like an escalator, the subtle displacement between any two sampled frames ($f_{i_1}, f_{i_2}, f_{i_3}$) is often lost, leading the model to perceive a series of static, nearly identical images rather than a continuous flow.}
    \label{fig:why_failed}
    \vskip -0.2in
\end{figure*}

In stark contrast, an MLLM processing the same video operates with a form of \textbf{frame-level myopia}. The dominant video understanding paradigm involves sampling discrete frames from the video at a certain rate and feeding these static images into the model for analysis. As illustrated in Figure~\ref{fig:why_failed}, when viewed in isolation, these frames are nearly indistinguishable. The subtle displacement of the escalator steps from one frame to the next is often smaller than the visual noise or may be entirely lost if the sampling rate is too low.

To illustrate this limitation, consider the stark contrast between this Escalator Problem and a high-signal motion event, such as a person walking across a room. In the latter case, even with sparse frame sampling, the significant change in the subject's silhouette, position, and the occlusion of background elements between frames provides a strong, unambiguous motion signal. The model can easily detect this change. The escalator, however, represents a near-worst-case scenario for this paradigm: its motion is characterized by uniform, continuous, and texturally repetitive flow. The visual evidence of motion is low-signal and distributed evenly across the entire object, rather than being concentrated in a distinct, moving entity. It is this specific failure to perceive motion in the absence of a high-signal, discrete moving object that defines the core of Implicit Motion Blindness.

The model, therefore, is presented with a series of snapshots that essentially say: ``This is an escalator,'' ``This is the same escalator,'' ``This is still the same escalator.''
Its training on vast datasets like ImageNet \cite{deng2009imagenet}, COCO \cite{lin2015coco}, or even large-scale video datasets has optimized it for semantic recognition—identifying objects, scenes, and discrete actions. However, it has not been sufficiently trained to understand the fine-grained physical processes that generate the visual data.
The model excels at answering ``What is this?'' but falters when asked, ``How is it behaving?''.
The critical directional signal, which is encoded in the temporal relationship between frames, is lost in the sampling process. This architectural blind spot is the root cause of its failure on the Escalator Test, as empirically demonstrated in Figure~\ref{fig:evidence_of_failure}, where state-of-the-art models fail to correctly describe the motion of two adjacent escalators.

\subsection{The Frame-Sampling Paradigm}

To deconstruct the ``\textit{frame-level myopia}'' with technical precision, we can formalize the general paradigm through which modern MLLMs process video. The core objective of such a model is to learn a conditional probability distribution $P(Y|V, T)$, which is the probability of generating a target text sequence $Y$ given an input video $V$ and a text prompt $T$.
This process involves four stages.

\paragraph{Symbol Definitions.}
\begin{description}
    \item[$V$] The raw input video, represented as a tensor $V \in \mathbb{R}^{T_v \times H \times W \times C}$, where $T_v$ is the total number of frames, $H, W$ are the frame dimensions, and $C$ is the number of color channels.
    \item[$T$] The input text prompt, a sequence of $M$ tokens $T = (w_1, w_2, \dots, w_M)$.
    \item[$Y$] The generated output text, a sequence of $L$ tokens $Y = (y_1, y_2, \dots, y_L)$.
    \item[$d$] The model's internal feature dimensionality.
\end{description}

\paragraph{Stage 1: Video Representation.}
This initial stage converts the continuous video signal into a discrete, ordered sequence of image patches. This is where the critical information loss occurs.

\noindent\textbf{Frame Sampling.} A sampling function, $\text{Sample}(\cdot)$, extracts $N$ frames from the full video $V$ to form a frame sequence $V_s$.
\begin{equation}
    V_s = \text{Sample}(V, N) = (f_1, f_2, \dots, f_N)
\end{equation}
where each frame $f_i \in \mathbb{R}^{H \times W \times C}$. Crucially, $N \ll T_v$.

\noindent\textbf{Patchification.} A patching function, $\text{Patch}(\cdot)$, divides each frame $f_i$ into $K$ smaller image patches.
\begin{equation}
    \text{Patch}(f_i) = (p_{i,1}, p_{i,2}, \dots, p_{i,K})
\end{equation}
where each patch $p_{i,j} \in \mathbb{R}^{P \times P \times C}$, and $P$ is the patch size.

\paragraph{Stage 2: Feature Encoding.}
This stage maps the discrete visual and textual data into a unified, high-dimensional continuous vector space.

\noindent\textbf{Visual Encoding.} A visual encoder $E_v$ (e.g., a Vision Transformer, ViT) transforms the sequence of all patches from all sampled frames into a sequence of $d$-dimensional visual features $Z_v$.
\begin{equation}
    Z_v = E_v(V_s) \in \mathbb{R}^{(N \times K) \times d}
\end{equation}

\noindent\textbf{Text Encoding.} A language encoder $E_l$ (e.g., a Transformer Encoder) converts the prompt $T$ into a sequence of $d$-dimensional text features $Z_l$.
\begin{equation}
    Z_l = E_l(T) \in \mathbb{R}^{M \times d}
\end{equation}

\paragraph{Stage 3: Multimodal Fusion.}
This core stage aligns and integrates information from the different modalities to form a unified, cross-modal understanding.

\noindent We define a generic fusion module $F_m$ that takes the independent visual and text feature sequences, $Z_v$ and $Z_l$, as input and outputs a fused hidden state sequence, $H_{\text{fused}}$.
\begin{equation}
    H_{\text{fused}} = F_m(Z_v, Z_l)
\end{equation}
Functionally, $F_m$ ``anchors'' textual concepts to visual evidence within a shared semantic space by modeling complex intra-modal and inter-modal relationships. This can be implemented via co-attention mechanisms, dedicated fusion layers, or by concatenating and passing features through a deep Transformer. For our purposes, we abstract away the specific implementation. The final output, $H_{\text{fused}} \in \mathbb{R}^{S \times d}$, is a context-aware representation ready for downstream tasks, where $S$ is the length of the fused sequence.

\paragraph{Stage 4: Autoregressive Generation.}
Based on the fused representation $H_{\text{fused}}$, an autoregressive decoder generates the final output text $Y$ token by token. Following the chain rule of probability, the generation process is formalized as:
\begin{equation}
    P(Y|V, T) = \prod_{t=1}^{L} P(y_t | y_{<t}, H_{\text{fused}})
\end{equation}
At each timestep $t$, the probability distribution for the next token is determined by a function $G$ (typically a linear layer followed a Softmax function):
\begin{equation}
    P(y_t | y_{<t}, H_{\text{fused}}) = G(H_{\text{fused}}, y_{<t})
\end{equation}
The model samples a token $y_t$ from this distribution, appends it to the input for the next step, and repeats the process until an end-of-sequence token is generated or a maximum length is reached.

This formalization makes the root of the Implicit Motion Blindness explicit: the irreversible loss of temporal continuity occurs at the very first step, in the $\text{Sample}(V, N)$ operation. No matter how sophisticated the subsequent fusion and generation stages are, they cannot recover the fine-grained motion information that was discarded before they ever saw the data.

\section{The Ripple Effect: Beyond Escalators, A Crisis of Trust}
\label{sec:ripple_effect}

\begin{table*}[ht]
    \centering
    \rowcolors{2}{gray!10}{white}
    \begin{tabularx}{\textwidth}{l X X}
        \toprule
        
        \rowcolor{gray!20}
        \textbf{Scenario} & \textbf{Challenge for AI (\textit{Implicit Motion})} & \textbf{Potential Consequence for BVI User} \\
        
        \midrule

        Escalator / Travelator & Perceiving direction of linear motion from a first-person perspective. & Wasted time, disorientation, social awkwardness, risk of falling. \\
        Revolving Door & Distinguishing between moving, stationary, or blocked states. & Attempting to enter a static or blocked door, risk of collision. \\
        Crowd Flow & Determining the dominant direction and speed of a moving crowd. & Inability to navigate efficiently, moving against the flow, social friction. \\
        Flowing Water / Puddles & Detecting subtle surface movement indicating flow or disturbance. & Risk of stepping into moving water (e.g., gutter, drain runoff), wet feet, potential slips. \\
        Automatic Sliding Doors & Perceiving the subtle initial movement indicating the doors are opening. & Hesitation, walking into a closed door, awkward waiting. \\
        Baggage Carousel & Identifying direction and speed of the conveyor belt. & Difficulty in positioning oneself to retrieve luggage, increased stress. \\
        
        \bottomrule
    \end{tabularx}
    \caption{A Spectrum of Implicit Motion Challenges for Assistive AI. The Escalator Problem is representative of a broader class of real-world scenarios where understanding implicit, continuous motion is critical for safe and effective navigation for BVI users. This table outlines several common situations where current AI approaches are likely to fail, highlighting the systemic nature of the issue.}
    \label{tab:spectrum}
\end{table*}

The inability of advanced AI to pass the \textit{Escalator Test} would be an academic curiosity if it were an isolated phenomenon.
However, its true significance lies in what it represents: a crack in the foundation of our approach to building assistive technologies for the real world.
This Implicit Motion Blindness is a systemic vulnerability that manifests across a wide spectrum of everyday scenarios.
The ultimate consequence of this vulnerability is not just functional failure, but the erosion of the single most critical element for any assistive tool: \textbf{user trust}.

\subsection{A Spectrum of Implicit Motion Challenges}

The Escalator Problem is a stand-in for a broader class of environmental interactions where perceiving subtle, continuous motion is not merely helpful, but essential for safety and social integration.
In each case, the core challenge remains the same: the critical information is encoded in the dynamic flow of the environment, not in the static identity of its components.
Consider the scenarios outlined in Table~\ref{tab:spectrum}. Navigating a public space like an airport or a busy street requires constant, implicit interpretation of motion cues that current MLLMs are ill-equipped to handle.

For a BVI user, an AI assistant that cannot distinguish the direction of a baggage carousel, the subtle sway of a revolving door, or the dominant flow of a crowd is not just inconvenient; it is functionally deficient.
It fails to provide the very layer of situational awareness that such a tool is meant to deliver.
The model might correctly identify a crowd of people, but it fails to provide the actionable insight: ``the crowd is moving towards your left.'' It might see a revolving door, but cannot answer the crucial question: ``is it safe to enter now?''. This gap between semantic description and perceptual understanding transforms the AI from a potential navigator into a mere commentator, narrating a scene without providing the means to act within it.

\subsection{The Consequence: A Crisis of Trust}

This leads to the most profound implication of our findings: a crisis of trust. An assistive tool that is predictably unreliable in common, everyday situations is worse than no tool at all. Trust in an assistive device is not built on its exceptional performance on benchmark tasks; it is forged through consistent, reliable behavior in the mundane, messy reality of daily life. When a user discovers their AI assistant cannot solve a problem as fundamental as the \textit{Escalator Test}, their confidence in the system's ability to handle more complex or higher-stakes situations—like identifying a safe gap in traffic to cross a street—is fundamentally undermined.

This issue speaks directly to the core challenges of \textbf{human-centered evaluation} and the \textbf{real-world deployment of AI-powered assistive systems}.
A system that is 99\% accurate on a dataset of discrete object labels but fails on 100\% of implicit motion tasks is, from a human-centered perspective, a brittle and therefore untrustworthy system.

This brittleness creates more than just functional unreliability; it imposes a significant \textbf{cognitive load} \cite{sweller1988cognitiveload} on the BVI user. In human-computer interaction, a primary goal of assistive technology is to reduce this load. However, a predictably unpredictable system forces the user into a constant state of vigilance, compelling them to second-guess the AI's output and determine which perceptions are reliable. This directly contradicts the technology's core purpose.
This transforms the interaction from a seamless assistive experience into a stressful cognitive puzzle, where the user must constantly deduce: ``Is this an object the AI is good at, or a motion it's blind to?''

Furthermore, we must distinguish between trust, a user's general belief in the system's competence and integrity, and reliance, the decision to depend on the system in a specific context. The Escalator Problem erodes fundamental trust at a systemic level. Consequently, even if the model performs perfectly on 99\% of static object recognition tasks, a rational user cannot rely on it for any task involving dynamic situational awareness. This transforms the AI from a potential partner into a mere ``situational commentator,'' one whose observations require constant, stressful verification, and whose failures risk not only physical safety but also social awkwardness and a diminished sense of independence.

Therefore, Implicit Motion Blindness is not simply a technical limitation to be incrementally improved. It is a fundamental barrier to establishing the user trust necessary for adoption and effective use. Without addressing this core perceptual deficit, we risk developing technologies that look impressive in demonstrations but fail silently and dangerously in the hands of the very people we aim to empower. The path forward requires us to acknowledge this crisis and re-evaluate the foundational principles upon which our assistive systems are built.
\section{A Call for a New Direction: From Recognition to Perception}
\label{sec:new_direction}

The discovery of \textit{motion blindness} necessitates more than an incremental fix; it demands a fundamental re-evaluation of our research trajectory in AI for accessibility. Continuing to scale models on ever-larger datasets of labeled objects and discrete actions will not solve the Escalator Problem. The solution lies in a paradigm shift: we must guide our models to evolve from engines of \textbf{recognition} to systems of \textbf{perception}.
A system of recognition asks, ``What is this object?''; a system of perception asks, ``How is this scene behaving, and what does that mean for me?''.
For an assistive AI, the latter is immeasurably more important.

\subsection{Move Beyond Frame-Based Semantics}

This paper's central position is that the computer vision community, particularly in its pursuit of assistive technologies, must consciously move beyond the limitations of semantic frame description and prioritize the development of true motion perception. The frame-sampling approach, while powerful for cataloging the contents of a video, is an abstraction that breaks the continuity of reality. It treats time as a slideshow, not a stream. This is a critical flaw when building tools designed to help a user navigate a dynamic, physical world.

We argue that Implicit Motion Blindness is a direct and inevitable consequence of this flawed abstraction. To build systems that are robust, trustworthy, and genuinely useful in real-world settings, we must treat motion not as an afterthought derived from comparing static frames, but as a primary and fundamental source of information. This requires a shift in architectural priorities, data collection philosophies, and the very definition of what it means for a model to understand a video.

\subsection{A Call for Human-Centered Benchmarks}

The emergence of real-time, conversational MLLMs makes the development of new evaluation paradigms not just an academic exercise, but a pressing safety imperative. The current landscape of benchmarks, while valuable for gauging progress in classification, detection, and semantic question-answering, is insufficient for evaluating real-world assistive performance.

Therefore, we issue a call for a community-wide effort to rethink evaluation itself. Table~\ref{tab:bench_comparison} provides a comparison that highlights the key differences between current benchmarks and our proposed human-centered approach.
We advocate for the development of a new class of evaluation protocols, co-designed with the BVI community, HCI experts, and accessibility professionals. These new benchmarks should:

\begin{itemize}
    \item \textbf{Prioritize Real-World Tasks:} Move beyond abstract VQA and incorporate tasks directly relevant to navigation and safety, such as those characterized by Implicit Motion Blindness, including determining escalator direction, crowd flow, and door state.
    \item \textbf{Measure Trustworthiness, Not Just Accuracy:} Develop metrics that capture system reliability and predictability. A model that correctly identifies an escalator's direction 100\% of the time is more valuable than a model that can name 10,000 objects but is unreliable on this critical task. This may involve measuring not just the correctness of an answer, but also its confidence, consistency, and its ability to signal uncertainty (i.e., to ``know what it doesn't know'').
    \item \textbf{Embrace Egocentric and Continuous Evaluation:} Shift from evaluating performance on curated, third-person video clips to assessing continuous performance in unstructured, egocentric first-person video streams, which more accurately reflect the input of a wearable assistive device.
\end{itemize}

\begin{table*}[ht]
    \centering
    \rowcolors{2}{gray!10}{white}
    \begin{tabularx}{\textwidth}{l X X}
        \toprule
        
        \rowcolor{gray!20}
        \textbf{Feature} & \textbf{Current Benchmarks} & \textbf{Proposed Human-Centered Benchmarks} \\
        
        \midrule
        
        Primary Task & Semantic Action Recognition & Fine-grained Physical Perception \\
        Motion Type & High-Signal / Discrete Actions & Low-Signal / Continuous Motion \\
        Static Bias & High (Keyframe-based) & Low (Requires Temporal Reasoning) \\
        Evaluation Metric & Classification Accuracy & Trust, Reliability, Safety \\
        Data Source & Curated Third-Person Video & Egocentric First-Person Streams \\
        
        \bottomrule
    \end{tabularx}
    \caption{A Comparison Between Current and Proposed Human-Centered Benchmarks. This table contrasts the focus of existing video understanding benchmarks with the proposed new paradigm. The shift moves from high-level action classification in curated videos towards evaluating trustworthiness and safety on continuous, first-person video streams, directly addressing the shortcomings that lead to Implicit Motion Blindness.}
    \label{tab:bench_comparison}
\end{table*}

\subsection{Promising Avenues for Exploration}

Solving this challenge will require innovation and an openness to exploring architectures and modalities beyond current MLLMs.
While we do not claim to have the solution, we encourage researchers to investigate several promising avenues that prioritize the physics of motion over the semantics of objects.

\paragraph{Hybrid Approaches}
Before the deep learning era, classical computer vision techniques like optical flow were purpose-built to estimate motion fields between frames.
Their efficacy for the Escalator Problem stems from a fundamentally different paradigm: instead of recognizing semantic objects, they directly compute pixel-level displacement vectors between frames, a process that is agnostic to what the object \textit{is}. This makes them inherently robust to challenges where MLLMs are confounded by the repetitive, semantically-identical appearance of the steps from one frame to the next.
We propose reintegrating these robust and efficient algorithms into modern architectures. Instead of being replaced by end-to-end models, optical flow could serve as a strong motion prior, a parallel processing stream, or a verification mechanism that grounds an MLLM's abstract understanding in the physical reality of pixel movement.

A potential implementation could be a two-stream fusion architecture. One stream would consist of the conventional MLLM, processing sparsely sampled frames to excel at semantic understanding—answering ``What is this object?''. In parallel, a second, lightweight stream dedicated to motion analysis, perhaps using a modern optical flow network like RAFT \cite{teed2020raft}, would process denser frame pairs to generate a robust motion vector field, answering ``How is this scene behaving?''. The outputs could be fused, where the motion stream acts as an attentional mechanism, guiding the MLLM to focus on dynamic regions, or as a verifier that fact-checks the MLLM's generated descriptions against physical motion evidence.

\paragraph{New Sensing Modalities}
The very sensor we use—the conventional frame-based camera—contributes to the problem. We should explore alternative sensors like \textbf{event cameras}. Unlike traditional cameras that capture frames at a fixed rate, event cameras are bio-inspired sensors that asynchronously report pixel-level brightness changes. They are inherently motion-detectors.
Their data is sparse, has high temporal resolution, and requires low power consumption, making them nearly ideal sensors for detecting subtle motion in a wearable, real-time assistive device.

Adopting these sensors, however, presents its own fundamental research challenges that the community must address. Event data, a sparse stream of asynchronous events, is incompatible with the dense, grid-based input expected by architectures like Vision Transformers. This necessitates a new line of inquiry: should we develop novel, non-Transformer architectures natively suited to processing such sparse event streams, or should we focus on creating sophisticated methods to ``render'' or ``tokenize'' event data into a dense, frame-like representation that existing MLLMs can comprehend without losing the crucial high-temporal-resolution information?

\paragraph{Physics-Informed Learning}
A more ambitious, long-term direction is to develop models with an innate, intuitive understanding of physics. This involves creating learning frameworks where models are not just trained on visual data, but are constrained or guided by the fundamental laws of motion, gravity, and continuity. A physics-informed model wouldn't need to have seen a million escalators; it would ``understand'' that a rigid set of stairs in continuous motion must be moving in a coherent direction.
\section{Limitations and Future Discourse}
\label{sec:limitations}

In advocating for this paradigm shift, we acknowledge several important considerations and potential counterarguments.

First, one might argue that advanced prompt engineering could mitigate this issue. While a carefully crafted prompt (e.g., ``Describe the motion of the steps relative to the stationary handrail'') might yield better results in isolated cases, we contend it is not a fundamental solution. This approach shifts the cognitive burden to the BVI user, requiring them to diagnose the AI's failure mode and formulate a precise query mid-navigation, which runs counter to the goal of a seamless assistive experience. More importantly, if the essential motion information is already lost in the initial $\text{Sample}(V, N)$ operation, no amount of clever prompting can fully recover it.

A second counterargument might point to alternative temporal processing pipelines, such as employing denser or adaptive frame sampling. While these methods may offer marginal improvements, they do not fundamentally resolve the Escalator Problem. For one, they often introduce significant computational overhead, a critical barrier for real-time applications on power-constrained wearable devices. More fundamentally, for textures as repetitive and uniform as escalator steps, even denser sampling can fail to provide a clear directional signal to a model optimized for semantic recognition. This suggests the issue is less about the \textit{quantity} of temporal data and more about the \textit{quality} and \textit{nature} of its representation. The problem lies not just in sampling, but in a paradigm that prioritizes \textit{what} an object is over \textit{how} it is behaving.

Furthermore, the call to prioritize physical perception should not be misinterpreted as a call to abandon semantic richness. The ideal future system is not one that sacrifices the MLLM's powerful descriptive abilities but one that gracefully integrates physical and semantic understanding. The key challenge for future research will be to resolve the potential trade-off between these two modes of perception, creating a model that both understands the physics of a scene and can describe it with human-like nuance.

Finally, while our position is that this problem is not merely a matter of scale, we acknowledge that the interplay between model size and emergent abilities is complex. Future work should critically examine whether simply scaling models further on undifferentiated data exacerbates this issue by reinforcing static biases, or if specific, physics-aware training regimes at scale can lead to a genuine, generalizable understanding of motion.

\section{Conclusion}
\label{sec:conclusion}

In this paper, we have illuminated a critical gap between the capabilities of modern AI and the demands of real-world assistive navigation. Through the simple, yet profound, \textit{Escalator Problem}, we demonstrated that state-of-the-art MLLMs suffer from \textit{Implicit Motion Blindness}—a fundamental inability to perceive low-signal, continuous motion. We have argued that this is not an isolated flaw, but an inevitable consequence of the dominant frame-sampling paradigm, which sacrifices the essence of temporal dynamics for computational efficiency. This limitation poses a significant barrier to building assistive tools that BVI users can truly trust with their safety and independence.

As a position paper, our primary objective was to diagnose this foundational issue and chart a new course for the community. We have formally identified and articulated the problem, analyzed how this technical deficit metastasizes into a crisis of user trust across a spectrum of everyday scenarios, and we have issued a clear call to action.
We contend that true progress in AI for accessibility cannot be measured by performance on static benchmarks alone. It requires a paradigm shift: from semantic \textbf{recognition} to holistic physical \textbf{perception}.

The path forward demands that we rethink our foundational assumptions.
We must develop human-centered benchmarks co-designed with the BVI community, explore hybrid architectures that integrate classical motion-aware techniques, and investigate novel sensing modalities such as event cameras. By redirecting our focus towards the physics of motion and the unwavering standard of user trust, we can begin to build the next generation of assistive AI—systems that do not just describe the world, but empower users to navigate it safely and confidently.
{
    \small
    \bibliographystyle{ieeenat_fullname}
    \bibliography{main}
}

\end{document}